%% file: 0-main.tex
\documentclass[11pt,a4paper]{article}
\usepackage{naacl2021}
\usepackage{hyperref}
\usepackage{times}
\usepackage{latexsym}
\usepackage[utf8]{inputenc}
\usepackage{multicol, latexsym, amsmath, amssymb}
\usepackage{blindtext}
\usepackage{subcaption}
\usepackage{caption}
\usepackage{multirow}
\usepackage{graphicx}
\usepackage{algorithm}
\usepackage{algorithmic}
\usepackage{comment}
\usepackage{float}
\usepackage{CJKutf8}

\usepackage{microtype}

\usepackage{array}
\newcolumntype{P}[1]{>{\centering\arraybackslash}p{#1}}
\newcolumntype{M}[1]{>{\centering\arraybackslash}m{#1}}

\newcommand{\ABC}{\textsc{ASAP}}

\title{ASAP: A Chinese Review Dataset Towards Aspect Category Sentiment Analysis and Rating Prediction}

\author{Jiahao Bu\textsuperscript{1}\thanks{\enspace Equal contribution.}, Lei Ren\textsuperscript{1}\footnotemark[1], Shuang Zheng\textsuperscript{1,2}\footnotemark[1], Yang Yang\textsuperscript{1},
Jingang Wang\textsuperscript{1}\thanks{\enspace Corresponding author.}, Fuzheng Zhang\textsuperscript{1}, Wei Wu\textsuperscript{1} \\
 \textsuperscript{1}Meituan, Beijing, China \\
 \textsuperscript{2}School of Economics and Management, Dalian University of Technology, Dalian, China \\ 
  \texttt{\{bujiahao,renlei04,zhengshuang04,yangyang113\}@meituan.com} \\ 
  \texttt{\{wangjingang02,zhangfuzheng,wuwei30\}@meituan.com} \\
  }

\date{}

\begin{document}
\maketitle


\begin{abstract}
Sentiment analysis has attracted increasing attention in e-commerce.
The sentiment polarities underlying user reviews are of great value for business intelligence.
Aspect category sentiment analysis (ACSA) and review rating prediction (RP) are two essential tasks to detect the fine-to-coarse sentiment polarities.
ACSA and RP are highly correlated and usually employed jointly in real-world e-commerce scenarios.
While most public datasets are constructed for ACSA and RP separately, which may limit the further exploitations of both tasks.
To address the problem and advance related researches, we present a large-scale Chinese restaurant review dataset \textbf{ASAP} including $46,730$ genuine reviews from a leading online-to-offline (O2O) e-commerce platform in China.
Besides a $5$-star scale rating, each review is manually annotated according to its sentiment polarities towards $18$ pre-defined aspect categories.
We hope the release of the dataset could shed some light on the field of sentiment analysis.
Moreover, we propose an intuitive yet effective joint model for ACSA and RP.
Experimental results demonstrate that the joint model outperforms state-of-the-art baselines on both tasks.
\end{abstract}

\input{1-Introduction}
\input{2-RelatedWork}
\input{3-Dataset}
\input{4-Methodology}
\input{5-Experiments}

\section{Conclusion}
This paper presents \textbf{ASAP}, a large-scale Chinese restaurant review dataset towards aspect category sentiment analysis (ACSA) and rating prediction (RP).
\textsc{ASAP} consists of $46,730$ restaurant user reviews with star ratings from a leading e-commerce platform in China.
Each review is manually annotated according to its sentiment polarities on $18$ fine-grained aspect categories.
Besides evaluations of ACSA and RP models on \textsc{ASAP} separately, we also propose a joint model to address ACSA and RP synthetically, which outperforms other state-of-the-art baselines considerably.
we hope the release of \textsc{ASAP} could push forward related researches and applications.

\clearpage
\bibliography{naacl2021}
\bibliographystyle{acl_natbib}


\end{document}

%% file: 1-Introduction.tex
\section{Introduction}
With the rapid development of e-commerce, massive user reviews available on e-commerce platforms are becoming valuable resources for both customers and merchants.
Aspect-based sentiment analysis(ABSA) on user reviews is a fundamental and challenging task which attracts interests from both academia and industries~\citep{hu2004mining,ganu2009beyond,jo2011aspect,kiritchenko2014nrc}.
According to whether the aspect terms are explicitly mentioned in texts, ABSA can be further classified into aspect term sentiment analysis (ATSA) and aspect category sentiment analysis (ACSA), we focus on the latter which is more widely used in industries.
Specifically, given a review "Although the fish is delicious, the waiter is horrible!", the ACSA task aims to infer the sentiment polarity over aspect category \textit{food} is positive while the opinion over the aspect category \textit{service} is negative. 

\begin{figure}[htb]
	\centering
	\includegraphics[height=0.25\textwidth]{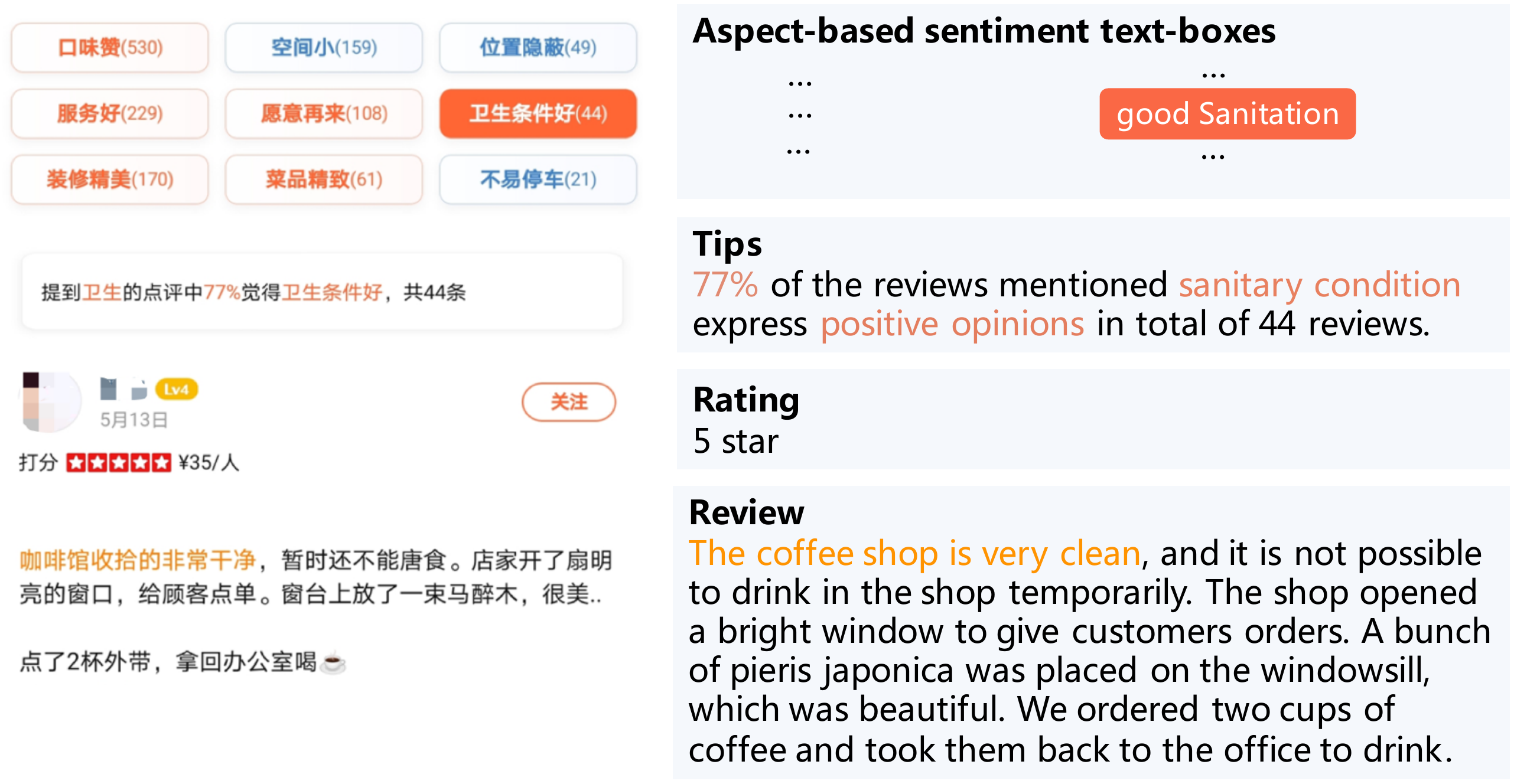}
	\caption{The user interface of a coffee shop on a popular e-commerce App. The top aspect-based sentiment text-boxes display aspect categories and sentiment polarities. The orange text-boxes are positive, while the blue ones are negative. The reviews mentioning the clicked aspect category (e.g., \textit{good sanitation}) with ratings are shown below. The text spans mentioning the aspect categories are also highlighted.}
	\label{fig:dianping}
\end{figure}

The user interfaces of e-commerce platforms are more intelligent than ever before with the help of ACSA techniques.
For example, \autoref{fig:dianping} presents the detail page of a coffee shop on a popular e-commerce platform in China.
The upper aspect-based sentiment text-boxes display the aspect categories (e.g., \textit{food}, \textit{sanitation}) mentioned frequently in user reviews and the aggregated sentiment polarities on these aspect categories (the orange ones represent positive and the blue ones represent negative).
Customers can focus on corresponding reviews effectively by clicking the aspect-based sentiment text-boxes they care about (e.g., the orange filled text-box ``\begin{CJK*}{UTF8}{gkai}卫生条件好\end{CJK*}'' (\textit{good sanitation})).
Our user survey based on $7,824$ valid questionnaires demonstrates that $80.08\%$ customers agree that the aspect-based sentiment text-boxes are helpful to their decision-making on restaurant choices.
Besides, the merchants can keep track of their cuisines and service qualities with the help of the aspect-based sentiment text-boxes.
Most Chinese e-commerce platforms such as Taobao\footnote{\url{https://www.taobao.com/}}, Dianping\footnote{\url{https://www.dianping.com/}}, and Koubei\footnote{\url{https://www.koubei.com/}} deploy the similar user interfaces to improve user experience. 

\begin{figure}[htb]
	\centering
	\includegraphics[width=0.45\textwidth]{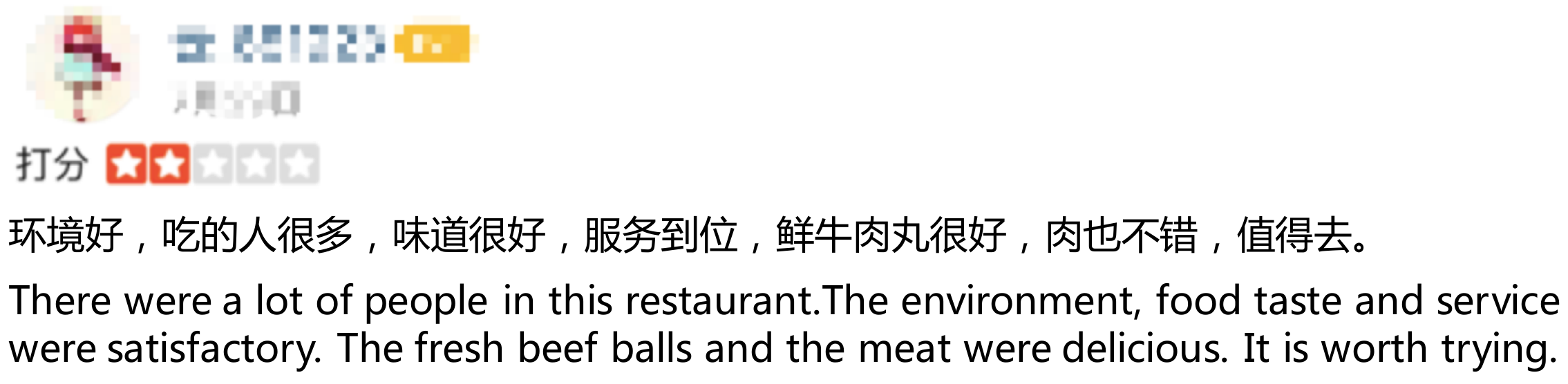}
	\caption{A content-rating disagreement case. The review holds a $2$-star rating while all the mentioned aspects are super positive.}
	\label{fig:review-rating}
\end{figure}

Users also publish their overall $5$-star scale ratings together with reviews. 
\autoref{fig:dianping} displays a sample of $5$-star rating to the coffee shop.
In comparison to fine-grained aspect sentiment, the overall review rating is usually a coarse-grained synthesis of the opinions on multiple aspects.
Rating prediction(RP)~\citep{jin2016jointly,li-etal-2018-document,wu2019arp} which aims to predict the “seeing stars” of reviews also has wide applications.
For example, to promise the aspect-based sentiment text-boxes accurate, unreliable reviews should be removed before ACSA algorithms are performed.
Given a piece of user review, we can predict a rating for it based on the overall sentiment polarity underlying the text.
We assume the predicted rating of the review should be consistent with its ground-truth rating as long as the review is reliable.
If the predicted rating and the user rating of a review disagree with each other explicitly, the reliability of the review is doubtful.
\autoref{fig:review-rating} demonstrates an example review of low-reliability.
In summary, RP can help merchants to detect unreliable reviews.

Therefore, both ACSA and RP are of great importance for business intelligence in e-commerce, and they are highly correlated and complementary.
ACSA focuses on predicting its underlying sentiment polarities on different aspect categories, while RP focuses on predicting the user's overall feelings from the review content.
We reckon these two tasks are highly correlated and better performance could be achieved by considering them jointly.

As far as we know, current public datasets are constructed for ACSA and RP separately, which limits further joint explorations of ACSA and RP.
To address the problem and advance the related researches, this paper presents a large-scale Chinese restaurant review dataset for \textbf{A}spect category \textbf{S}entiment \textbf{A}nalysis and rating \textbf{P}rediction, denotes as \textbf{ASAP} for short. 
All the reviews in ASAP are collected from the aforementioned e-commerce platform.
There are $46,730$ restaurant reviews attached with $5$-star scale ratings.
Each review is manually annotated according to its sentiment polarities towards $18$ fine-grained aspect categories.
To the best of our knowledge, ASAP is the largest Chinese large-scale review dataset towards both ACSA and RP tasks.

We implement several state-of-the-art (SOTA) baselines for ACSA and RP and evaluate their performance on ASAP.
To make a fair comparison, we also perform ACSA experiments on a widely used SemEval-2014 restaurant review dataset~\citep{SemEval2014}.
Since BERT~\citep{Devlin2018BERTPO} has achieved great success in several natural language understanding tasks including sentiment analysis~\citep{xu2019bert,sun2019utilizing,mams}, we propose a joint model that employs the fine-to-coarse semantic capability of BERT.
Our joint model outperforms the competing baselines on both tasks.

Our main contributions can be summarized as follows. 
(1) We present a large-scale Chinese review dataset towards aspect category sentiment analysis and rating prediction, named as \textsc{ASAP}, including as many as $46,730$ real-world restaurant reviews annotated from $18$ pre-defined aspect categories. Our dataset has been released at \url{https://github.com/Meituan-Dianping/asap}. 
(2) We explore the performance of widely used models for ACSA and RP on \textsc{ASAP}.
(3) We propose a joint learning model for ACSA and RP tasks. Our model achieves the best results both on \textsc{ASAP} and SemEval \textsc{Restaurant} datasets. 

%% file: 2-RelatedWork.tex
\section{Related Work and Datasets}
\textbf{Aspect Category Sentiment Analysis.} 
ACSA~\citep{Zhou2015Representation,Movahedi2019Aspect,Ruder2016A,can} aims to predict sentiment polarities on all aspect categories mentioned in the text.
The series of SemEval datasets consisting of user reviews from e-commerce websites have been widely used and pushed forward related research~\citep{ATAE,IAN,xu2019bert,sun2019utilizing,mams}.
The SemEval-2014 task-4 dataset (SE-ABSA14)~\citep{SemEval2014} is composed of laptop and restaurant reviews. 
The restaurant subset includes $5$ aspect categories (i.e., \textit{Food}, \textit{Service}, \textit{Price}, \textit{Ambience} and \textit{Anecdotes/Miscellaneous}) and $4$ polarity labels (i.e., \textit{Positive}, \textit{Negative}, \textit{Conflict} and \textit{Neutral}).
The laptop subset is not suitable for ACSA.
The SemEval-2015 task-12 dataset (SE-ABSA15)~\citep{SemEval2015} builds upon SE-ABSA14 and defines its aspect category as a combination of an entity type and an attribute type(e.g., \textit{Food\#Style\_Options}).
The SemEval-2016 task-5 dataset (SE-ABSA16)~\citep{SemEval2016} extends SE-ABSA15 to new domains and new languages other than English.
MAMS~\citep{mams} tailors SE-ABSA14 to make it more challenging, in which each sentence contains at least two aspects with different sentiment polarities.

Compared with the prosperity of English resources, high-quality Chinese datasets are not rich enough. 
``ChnSentiCorp''~\citep{tan2008empirical}, ``IT168TEST''~\citep{zagibalov2008unsupervised}, ``Weibo''\footnote{\url{http://tcci.ccf.org.cn/conference/2014/pages/page04\_dg.html}}, ``CTB''~\citep{li2014recursive} are $4$ popular Chinese datasets for general sentiment analysis.
However, aspect category information is not annotated in these datasets.
~\citet{zhao2014creating} presents two Chinese ABSA datasets for consumer electronics (mobile phones and cameras).
Nevertheless, the two datasets only contain $400$ documents ($\sim4000$ sentences), in which each sentence only mentions one aspect category at most. 
BDCI\footnote{\url{https://www.datafountain.cn/competitions/310}} automobile opinion mining and sentiment analysis dataset~\citep{dai2019multi} contains $8, 290$ user reviews in automobile industry with $10$ pre-defined categories. 
~\citet{peng2017review} summarizes available Chinese ABSA datasets.
While most of them are constructed through rule-based or machine learning-based approaches, which inevitably introduce additional noise into the datasets.
Our ASAP excels above Chinese datasets both on quantity and quality.

\noindent \textbf{Rating Prediction.} Rating prediction (RP) aims to predict the “seeing stars” of reviews, which represent the overall ratings of reviews. 
In comparison to fine-grained aspect sentiment, the overall review rating is usually a coarse-grained synthesis of the opinions on multiple aspects.
~\citet{ganu2009beyond, li2011incorporating,chen2018neural} form this task as a text classification or regression problem.
Considering the importance of opinions on multiple aspects in reviews, recent years have seen numerous work~\citep{jin2016jointly, cheng2018aspect,li-etal-2018-document,wu2019arp} utilizing the information of the aspects to improve the rating prediction performance.
This trending also inspires the motivation of ASAP.

Most RP datasets are crawled from real-world review websites and created for RP specifically.
Amazon Product Review English dataset~\citep{mcauley2013hidden} containing product reviews and metadata from Amazon has been widely used for RP~\citep{cheng2018aspect,mcauley2013hidden}. 
Another popular English dataset comes from Yelp Dataset Challenge 2017\footnote{\url{http://www.yelp.com/dataset\_challenge/}}, which includes reviews of local businesses in $12$ metropolitan areas across $4$ countries.
Openrice\footnote{\url{https://www.openrice.com}} is a Chinese RP dataset composed of $168,142$ reviews.
Both the English and Chinese datasets don't annotate fine-grained aspect category sentiment polarities.


%% file: 3-Dataset.tex
\section{Dataset Collection and Analysis}
\label{sec:dataset}
\subsection{Data Construction \& Curation}
We collect reviews from one of the most popular O2O e-commerce platforms in China, which allows users to publish coarse-grained star ratings and writing fine-grained reviews to restaurants (or places of interest) they have visited. 
In the reviews, users comment on multiple aspects either explicitly or implicitly, including \textit{ambience},\textit{price}, \textit{food}, \textit{service}, and so on.

First, we retrieve a large volume of user reviews from popular restaurants holding more than $50$ user reviews randomly.
Then, $4$ pre-processing steps are performed to promise the ethics, quality, and reliability of the reviews. 
(1) User information (e.g., user-ids, usernames, avatars, and post-times) are removed due to privacy considerations. 
(2) Short reviews with less than $50$ Chinese characters, as well as lengthy reviews with more than $1000$ Chinese characters are filtered out.
(3) If the ratio of non-Chinese characters within a review is over $70$\%, the review is discarded.
(4) To detect the low-quality reviews (e.g., advertising texts), we build a BERT-based classifier with an accuracy of $97\%$ in a leave-out test-set. The reviews detected as low-quality by the classifier are discarded too. 

\subsection{Aspect Categories}
Since the reviews already hold users' star ratings, this section mainly introduces our annotation details for ACSA.
In SE-ABSA14 restaurant dataset (denoted as \textsc{Restaurant} for simplicity), there are $5$ coarse-grained aspect categories, including \textit{food}, \textit{service}, \textit{price}, \textit{ambience} and \textit{miscellaneous}.
After an in-depth analysis of the collected reviews, we find the aspect categories mentioned by users are rather diverse and fine-grained.
Take the text  ``...The restaurant holds a high-end decoration but is quite noisy since a wedding ceremony was being held in the main hall... (\begin{CJK*}{UTF8}{gkai}...环境看起来很高大上的样子，但是因为主厅在举办婚礼非常混乱，感觉特别吵...\end{CJK*})'' in \autoref{tb:dataset} for example, 
the reviewer actually expresses opposite sentiment polarities on two fine-grained aspect categories related to \textit{ambience}.
The restaurant's decoration is very high-end (\textit{Positive}), while it's very noisy due to an ongoing ceremony (\textit{Negative}).
Therefore, we summarize the frequently mentioned aspects and refine the $5$ coarse-grained categories into $18$ fine-grained categories.
We replace \textit{miscellaneous} with \textit{location} since we find users usually review the restaurants' location (e.g., whether the restaurant is easy to reach by public transportation.).
We denote the aspect category as the form of ``\textit{Coarse-grained Category\#Fine-grained Categoty}'', such as ``\textit{Food\#Taste}'' and  ``\textit{Ambience\#Decoration}''.
The full list of aspect categories and definitions are listed in \autoref{tb:full-list}.

\begin{table*}[htb]
   \small
   \center
   \caption{The full list of $18$ aspect categories and definitions.}
   \begin{tabular}{p{0.25\textwidth}p{0.20\textwidth}p{0.25\textwidth}p{0.20\textwidth}}
   \hline
   Aspect category & Definition & Aspect category & Definition \\ \hline
   \parbox{0.25\textwidth}{\textit{Food\#Taste}\\ (\begin{CJK*}{UTF8}{gkai}口味\end{CJK*})} & Food taste & \parbox{0.25\textwidth}{\textit{Location\#Easy\_to\_find}\\(\begin{CJK*}{UTF8}{gkai}是否容易寻找\end{CJK*})}  & Whether the restaurant is easy to find  \\
   \parbox{0.25\textwidth}{\textit{Food\#Appearance}\\(\begin{CJK*}{UTF8}{gkai}外观\end{CJK*})} & Food appearance & \parbox{0.25\textwidth}{\textit{Service\#Queue}\\(\begin{CJK*}{UTF8}{gkai}排队时间\end{CJK*})}  &  Whether the queue time is acceptable \\
   \parbox{0.25\textwidth}{\textit{Food\#Portion}\\ (\begin{CJK*}{UTF8}{gkai}分量\end{CJK*})} & Food portion &  \parbox{0.25\textwidth}{\textit{Service\#Hospitality}\\ (\begin{CJK*}{UTF8}{gkai}服务人员态度\end{CJK*})} & Waiters/waitresses' attitude/hospitality  \\  
   \parbox{0.25\textwidth}{\textit{Food\#Recommend}\\(\begin{CJK*}{UTF8}{gkai}推荐程度\end{CJK*})} & Whether the food is worth being recommended &  \parbox{0.25\textwidth}{\textit{Service\#Parking}\\(\begin{CJK*}{UTF8}{gkai}停车方便\end{CJK*})} & Parking convenience  \\ 
   \parbox{0.25\textwidth}{\textit{Price\#Level}\\(\begin{CJK*}{UTF8}{gkai}价格水平\end{CJK*})} & Price level &  \parbox{0.25\textwidth}{\textit{Service\#Timely}\\(\begin{CJK*}{UTF8}{gkai}点菜/上菜速度\end{CJK*})} & Order/Serving time \\
   \parbox{0.25\textwidth}{\textit{Price\#Cost\_effective}\\(\begin{CJK*}{UTF8}{gkai}性价比\end{CJK*})} & Whether the restaurant is cost-effective & \parbox{0.25\textwidth}{\textit{Ambience\#Decoration}\\(\begin{CJK*}{UTF8}{gkai}装修\end{CJK*})} & Decoration level \\
   \parbox{0.25\textwidth}{\textit{Price\#Discount}\\(\begin{CJK*}{UTF8}{gkai}折扣力度\end{CJK*})} & Discount strength  & \parbox{0.25\textwidth}{\textit{Ambience\#Noise}\\(\begin{CJK*}{UTF8}{gkai}嘈杂情况\end{CJK*})} & Whether the restaurant is noisy   \\
   \parbox{0.25\textwidth}{\textit{Location\#Downtown}\\(\begin{CJK*}{UTF8}{gkai}位于商圈附近\end{CJK*})} & Whether the restaurant is located near downtown  &  \parbox{0.25\textwidth}{\textit{Ambience\#Space}\\ (\begin{CJK*}{UTF8}{gkai}就餐空间\end{CJK*})} & Dining Space and Seat Size \\
   \parbox{0.25\textwidth}{\textit{Location\#Transportation}\\(\begin{CJK*}{UTF8}{gkai}交通方便\end{CJK*})} & Convenient public transportation to the restaurant &  \parbox{0.25\textwidth}{\textit{Ambience\#Sanitary}\\(\begin{CJK*}{UTF8}{gkai}卫生情况\end{CJK*})} & Sanitary condition \\
   	\hline
  \end{tabular}
   \label{tb:full-list}
\end{table*}

\subsection{Annotation Guidelines \& Process}
Bearing in mind the pre-defined $18$ aspects, assessors are asked to annotate sentiment polarities towards the mentioned aspect categories of each review. 
Given a review, when an aspect category is mentioned within the review either explicitly and implicitly, the sentiment polarity over the aspect category is labeled as $1$ (\textit{Positive}), $0$ (\textit{Neutral}) or $-1$ (\textit{Negative}) as shown in \autoref{tb:dataset}.

We hire $20$ vendor assessors, $2$ project managers, and $1$ expert reviewer to perform annotations.
Each assessor needs to attend a training to ensure their intact understanding of the annotation guidelines.
Three rounds of annotation are conducted sequentially. 
First, we randomly split the whole dataset into $10$ groups, and every group is assigned to $2$ assessors to annotate independently. 
Second, each group is split into $2$ subsets according to the annotation results, denoted as \textbf{Sub-Agree} and \textbf{Sub-Disagree}. \textbf{Sub-Agree} comprises the data examples with agreement annotation, and \textbf{Sub-Disagree} comprises the data examples with disagreement annotation.
\textbf{Sub-Agree} will be reviewed by assessors from other groups.
The controversial examples during the review are considered as difficult cases.
\textbf{Sub-Disagree} will be reviewed by the $2$ project managers independently and then discuss to reach an agreement annotation. 
The examples that could not be addressed after discussions are also considered as difficult cases.
Third, for each group, the difficult examples from two subsets are delivered to the expert reviewer to make a final decision.
More details of difficult cases and annotation guidelines during annotation are demonstrated in \autoref{tb:difficult-cases}.
\begin{table*}[htb]
   \small
  \centering
   \caption{Difficult cases and annotation guidelines.}
   \begin{tabular}{P{0.15\textwidth}P{0.2\textwidth}P{0.2\textwidth}P{0.2\textwidth}P{0.15\textwidth}}
   	\hline
   	Category & Example & Example (Translation) & Guideline & Annotation \\ \hline
    Change of sentiment over time  &  \begin{CJK*}{UTF8}{gkai}我之前挺喜欢这家餐厅的饭菜，不过今天的饭菜可不怎么样.\end{CJK*}  & I used to like the food of this restaurant, but the taste is not as expected today. & When there existed a sentiment drifting over time in the review, the most recent sentiment polarity is adopted. & (\textit{Food\#Taste}, $-1$)  \\
    Implicit sentiment polarity & \begin{CJK*}{UTF8}{gkai}比五星级酒店的餐厅差远了，而且在五星级酒店中餐厅里两个人吃一顿也就$500$左右就够了\end{CJK*}  & The restaurant was far worse than the dinning hall of any five-star hotel, considering that the meal for two people only cost $500$ CNY in a five-star hotel.  & Some reviewers express their polarities in an implicit manner instead of expressing their feelings directly. The implicit sentiment polarity is adopted to complete the annotation. & (\textit{Price\#Level}, $-1$) \\
    Conflict opinions & \begin{CJK*}{UTF8}{gkai}这道菜有点咸，但是味道很赞。\end{CJK*} & This dish was a bit salty, but it tasted great. & When there existed multiple sentiment polarities toward the same aspect-category, the dominant sentiment is chosen. & (\textit{Food\#Taste}, $1$)   \\
    Mild sentiment & \begin{CJK*}{UTF8}{gkai} 饭菜还可以，不过也算不上特别好吃。\end{CJK*} & The food was okay, but nothing great. & The “neutral” label applies to mildly positive or mildly negative sentiment & (\textit{Food\#Taste}, $0$)   \\
    Irrelevant restaurant & \begin{CJK*}{UTF8}{gkai} 上次去的一家店很难吃，今天来了这家新的，感觉很好吃。\end{CJK*} & The food of the shop which I went to last time was very bad. Today I came to this new one. I felt very good. & The review mentions restaurant that the user has visited in the past. We only focus on the restaurant being reviewed  & (\textit{Food\#Taste}, $1$)   \\
   	\hline
   \end{tabular}
   \label{tb:difficult-cases}
\end{table*}

Finally, \textsc{ASAP} corpus consists of $46,730$ pieces of real-world user reviews, and we split it into a training set ($36,850$), a validation set ($4,940$) and a test set ($4,940$) randomly.
\autoref{tb:dataset} presents an example review of \textsc{ASAP} and corresponding annotations on the $18$ aspect categories.

  \begin{table*}[htb]
  \renewcommand\arraystretch{3.8}
 	\small
 	\centering
 	\caption{A review example in \textsc{ASAP}, with overall star rating and aspect category sentiment polarity annotations.}\label{tb:dataset}
 	\begin{tabular}{M{0.42\textwidth}cM{0.15\textwidth}cM{0.15\textwidth}c}
 		\hline
 		Review  & Rating & Aspect Category & Label & Aspect Category & Label \\ \hline
 		\multirow{9}{*}{\parbox{0.42\textwidth}{With convenient traffic, the restaurant holds a high-end decoration, but quite noisy because a wedding ceremony was being held in the main hall. Impressed by its delicate decoration and grand appearance though, we had to wait for a while at the weekend time. However, considering its high price level, the taste is unexpected. We ordered the Kung Pao Prawn, the taste was acceptable and the serving size is enough, but the shrimp is not fresh. In terms of service, you could not expect too much due to the massive customers there. By the way, the free-served fruit cup was nice. Generally speaking, it was a typical wedding banquet restaurant rather than a comfortable place to date with friends. \\ \\  \begin{CJK*}{UTF8}{gkai} 交通还挺方便的，环境看起来很高大上的样子，但是因为主厅在举办婚礼非常混乱，特别吵感觉，但是装修的还不错，感觉很精致的装修，门面很气派，周末去的时候还需要等位。味道的话我觉得还可以但是跟价格比起来就很一般了，性价比挺低的，为了去吃宫保虾球的，但是我觉得也就那样吧虾不是特别新鲜，不过虾球很大，味道还行。服务的话由于人很多所以也顾不过来上菜的速度不快，但是有送水果杯还挺好吃的。总之就是典型的婚宴餐厅不是适合普通朋友吃饭的地方了。\end{CJK*} }} 
 		& \multirow{9}{*}{$3$-Star} & \parbox{0.15\textwidth}{\textit{Location\#Transportation}\\(\begin{CJK*}{UTF8}{gkai}交通方便\end{CJK*})} & $1$   & \parbox{0.15\textwidth}{\textit{Price\#Discount}\\ (\begin{CJK*}{UTF8}{gkai}折扣力度\end{CJK*}) }          & -     \\  
 		& & \parbox{0.15\textwidth}{\textit{Location\#Downtown} \\ (\begin{CJK*}{UTF8}{gkai}位于商圈附近\end{CJK*}) }          & - & \parbox{0.15\textwidth}{\textit{Ambience\#Decoration}\\ (\begin{CJK*}{UTF8}{gkai}装修\end{CJK*})} & $1$  \\
 		
 		& & \parbox{0.15\textwidth}{\textit{Location\#Easy\_to\_find}\\ (\begin{CJK*}{UTF8}{gkai}是否容易寻找\end{CJK*} )} & - & \parbox{0.15\textwidth}{\textit{Ambience\#Noise}\\ (\begin{CJK*}{UTF8}{gkai}嘈杂情况\end{CJK*})}  & $-1$ \\
 		
 		& & \parbox{0.15\textwidth}{\textit{Service\#Queue}\\ (\begin{CJK*}{UTF8}{gkai}排队时间\end{CJK*})} & - & 
 		\parbox{0.15\textwidth}{\textit{Ambience\#Space}\\ (\begin{CJK*}{UTF8}{gkai}就餐空间\end{CJK*})} & $1$ \\
 		
 		& & \parbox{0.15\textwidth}{\textit{Service\#Hospitality}\\ (\begin{CJK*}{UTF8}{gkai}服务人员态度\end{CJK*})}  & -     & \parbox{0.15\textwidth}{\textit{Ambience\#Sanitary}\\ (\begin{CJK*}{UTF8}{gkai}卫生情况\end{CJK*})} & - \\
 		
    	& & \parbox{0.15\textwidth}{\textit{Service\#Parking}\\ (\begin{CJK*}{UTF8}{gkai}停车方便\end{CJK*})}  & -     & \parbox{0.15\textwidth}{\textit{Food\#Portion}\\ (\begin{CJK*}{UTF8}{gkai}分量\end{CJK*})}  & $1$ \\
    	
		& & \parbox{0.15\textwidth}{\textit{Service\#Timely}\\ (\begin{CJK*}{UTF8}{gkai}点菜/上菜速度\end{CJK*})} & $-1$    & \parbox{0.15\textwidth}{\textit{Food\#Taste}\\ (\begin{CJK*}{UTF8}{gkai}口味\end{CJK*})}  & $1$ \\
		
    	& & \parbox{0.15\textwidth}{\textit{Price\#Level}\\(\begin{CJK*}{UTF8}{gkai}价格水平\end{CJK*})} & $0$     & \parbox{0.15\textwidth}{\textit{Food\#Appearance}\\(\begin{CJK*}{UTF8}{gkai}外观\end{CJK*})}  & - \\
    	
    	& & \parbox{0.15\textwidth}{\textit{Price\#Cost\_effective}\\ (\begin{CJK*}{UTF8}{gkai}性价比\end{CJK*})} & $-1$    & \parbox{0.15\textwidth}{\textit{Food\#Recommend}\\ (\begin{CJK*}{UTF8}{gkai}推荐程度\end{CJK*})} & - \\
    	\hline
 	\end{tabular}
 \label{tb:dataset}
 \end{table*} 

\subsection{Dataset Analysis}
\begin{figure*}[!t]
	\centering
	\vspace{-0.3cm}
	\includegraphics[width=0.95\linewidth]{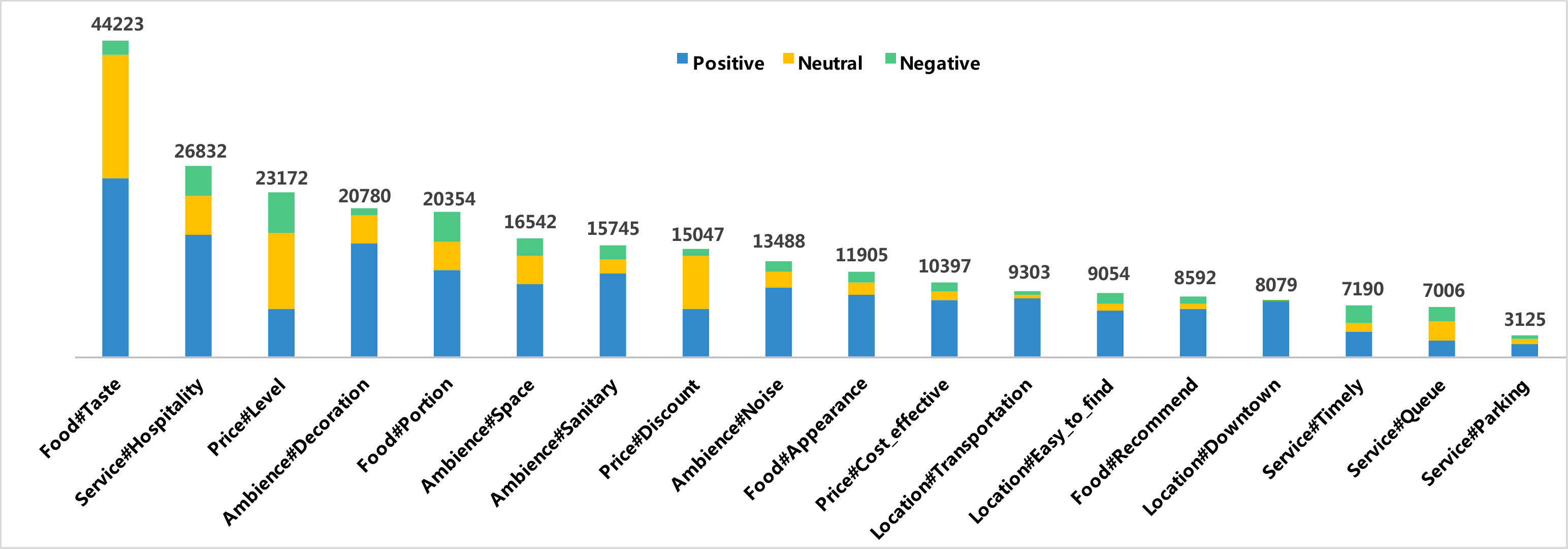}
	\vspace{-0.2cm}
	\caption{The distribution of $18$ fine-grained aspect categories in \textsc{ASAP}.}
	\label{fig:cate_dist}
\end{figure*}

\autoref{fig:cate_dist} presents the distribution of $18$ aspect categories in \textsc{ASAP}.
Because \textsc{ASAP} concentrates on the domain of restaurant, $94.7$\% reviews mention \textit{Food\#Taste} as expected. 
Users also pay great attention to aspect categories such as \textit{Service\#Hospitality}, \textit{Price\#Level} and \textit{Ambience\#Decoration}.
The distribution proves the advantages of \textsc{ASAP}, as users' fine-grained preferences could reflect the pros and cons of restaurants more precisely.

The statistics of \textsc{ASAP} are presented in \autoref{tb:distribution}.
We also include a tailored SE-ABSA14 \textsc{Restaurant} dataset for reference.
Please note that we remove the reviews holding aspect categories with sentiment polarity of ``conflict'' from the original \textsc{Restaurant} dataset. 

\begin{table*}[htb]
   \small
   \center
   \caption{The statistics and label/rating distribution of \textsc{ASAP} and \textsc{Restaurant}. The review length are counted by Chinese characters and English words respectively. The sentences are segmented with periods in \ABC{}, while \textsc{Restaurant} is a sentence-level dataset.}
   \scalebox{0.75}{
   \begin{tabular}{M{0.1\textwidth}ccM{0.1\textwidth}M{0.1\textwidth}M{0.05\textwidth}M{0.05\textwidth}ccccccc}
   	\hline
   	Dataset & Split & Reviews &  \parbox{0.1\textwidth}{\centering Average \\sentences\\per review} & \parbox{0.1\textwidth}{\centering Average\\aspects\\ per review} & \parbox{0.05\textwidth}{\centering Average\\length}  & Positive & Negative & Neutral & $1$-star & $2$-star & $3$-star & $4$-star & $5$-star \\ \hline
   	\multirow{3}{*}{\textsc{ASAP}} & Train & $36,850$ & $8.6$ & $5.8$ & $319.7$  & $133,721$ & $27,425$ & $52,225$ 
   	& $1,219$ & $1,258$ & $5,241$ & $13,362$ & $15,770$ \\
   	& Dev & $4,940$ & $8.7$ & $5.9$ & $319.9$ & $18,176$  & $3,733$  & $7,192$ 
   	& $151$ & $166$ & $784$ & $1,734$ & $2,105$  \\
   	& Test & $4,940$ & $8.3$ & $5.7$ & $317.1$ & $17,523$  & $3,813$ & $7,026$ 
   	& $165$ & $173$ & $717$ & $1,867$ & $2,018$ \\
   	\hline
   	\multirow{2}{*}{\textsc{Restaurant}} & Train & $2,855$ & $1$ & $1.2$ & $15.2$ & $2150$  & $822$& $498$  & - & - & - & - & -  \\
   	& Test & $749$ & $1$ & $1.3$ & $15.6$ &$645$  & $215$  & $94$ & - & - & - & - & -   \\
   	\hline
   \end{tabular}}
   \label{tb:distribution}
\end{table*}

Compared with \textsc{Restaurant}, \textsc{ASAP} excels in the quantities of training instances, which supports the exploration of recent data-intensive deep neural models.
\textsc{ASAP} is a review-level dataset, while \textsc{Restaurant} is a sentence-level dataset. 
The average length of reviews in \textsc{ASAP} is much longer, thus the reviews tend to contain richer aspect information. 
In \textsc{ASAP}, the reviews contain $5.8$ aspect categories in average, which is $4.7$ times of \textsc{Restaurant}.
Both review-level ACSA and RP are more challenging than their sentence-level counterparts.
Take the review in \autoref{tb:dataset} for example, the review contains several sentiment polarities towards multiple aspect categories.
In addition to aspect category sentiment annotations, \textsc{ASAP} also includes overall user ratings for reviews. 
With the help of \textsc{ASAP}, ACSA and RP can be further optimized either separately or jointly.

%% file: 4-Methodology.tex
\section{Methodology}
\label{sec:method}
\subsection{Problem Formulation}
We use $D$ to denote the collection of user review corpus in the training data. 
Given a review $R$ which consists of a series of words: $\{w_1,w_2,...,w_Z\}$, ACSA aims to predict the sentiment polarity $y_i \in \{Positive, Neutral, Negative\}$ of review $R$ with respect to the mentioned aspect category $a_i$, $i \in \{1,2,..., N\}$. 
$Z$ denotes the length of review $R$. 
$N$ is the number of pre-defined aspect categories (i.e., $18$ in this paper). 
Suppose there are $K$ mentioned aspect categories in $R$.
We define a mask vector $[p_1,p_2,...,p_N]$ to indicate the occurrence of aspect categories. 
When the aspect category $a_i$ is mentioned in $R$, $p_i = 1$, otherwise $p_i = 0$. 
So we have $\sum_{i=1}^N p_i = K$.
In terms of RP, it aims to predict the $5$-star rating score of $g$, which represents the overall rating of the given review $R$.

\subsection{Joint Model}
Given a user review, ACSA focuses on predicting its underlying sentiment polarities on different aspect categories, while RP focuses on predicting the user's overall feelings from the review content.
We reckon these two tasks are highly correlated and better performance could be achieved by considering them jointly.

\begin{figure}[!t]
	\centering
	\includegraphics[width=0.85\linewidth]{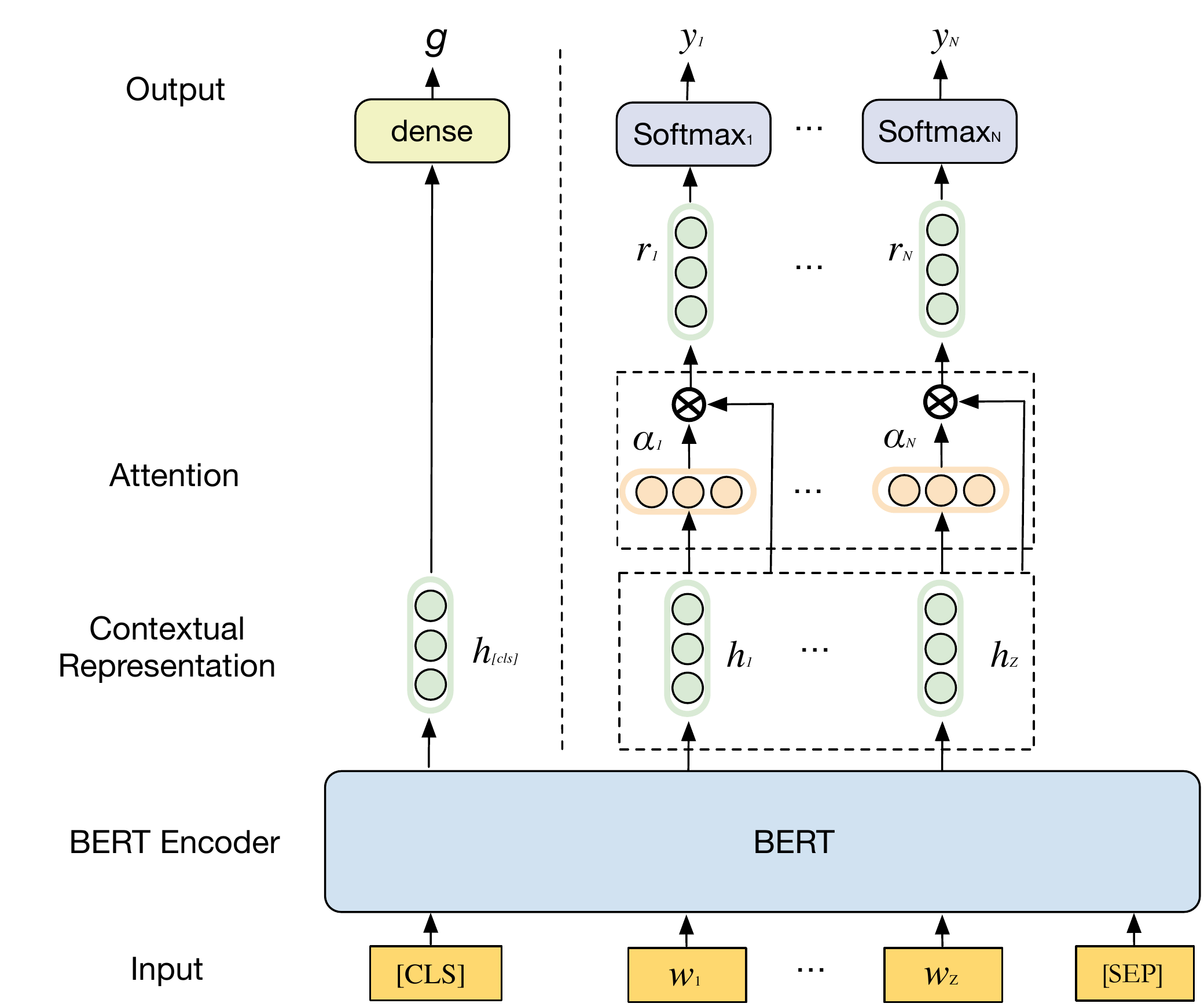}
	\caption{The framework of the proposed joint learning model. The right part of the dotted vertical line is used to predict multiple aspect category sentiment polarities, while the left part is used to predict the review rating.}
	\label{fig:framework}
\end{figure}

The advent of BERT has established the success of the ``pre-training and then fine-tuning'' paradigm for NLP tasks.
BERT-based models have achieved impressive results in ACSA~\citep{xu2019bert,sun2019utilizing,mams}.
Review rating prediction can be deemed as a single-sentence classification (regression) task, which could also be addressed with BERT.
Therefore, we propose a joint learning model to address ACSA and RP in a multi-task learning manner.
Our joint model employs the fine-to-coarse semantic representation capability of the BERT encoder.
\autoref{fig:framework} illustrates the framework of our joint model. 

\noindent\textbf{ACSA}
As shown in \autoref{fig:framework}, the token embeddings of the input review are generated through a shared BERT encoder.
Briefly, let $H \in \mathbb{R}^{d*Z}$ be the matrix consisting of token embedding vectors $\{h_1,...,h_Z\}$ that BERT produces, where $d$ is the size of hidden layers and $Z$ is the length of the given review. 
Since different aspect category information is dispersed across the content of $R$, we add an attention-pooling layer~\citep{ATAE} to aggregate the related token embeddings dynamically for every aspect category.
The attention-pooling layer helps the model focus on the tokens most related to the target aspect categories.
\begin{equation}
M_i^a = \tanh(W_i^a*H)
\end{equation}
\begin{equation}
\alpha_i = softmax(\omega_i^T*M_i^a)
\end{equation}
\begin{equation}
r_i = \tanh(W_i^{p}*H*\alpha_i^T)
\end{equation}
Where $W_i^a \in \mathbb{R}^{d*d}$, $M_i^a \in \mathbb{R}^{d*Z}$, $\omega_i \in \mathbb{R}^d$, $\alpha_i \in \mathbb{R}^Z, W_i^{p} \in \mathbb{R}^{d*d}$, and $r_i \in \mathbb{R}^d$.
$\alpha_i$ is a vector consisting of attention weights of all tokens which can selectively attend the regions of the aspect category related tokens, and $r_i$ is the attentive representation of review with respect to the $i_{th}$ aspect category $a_i$, $i \in \{1,2,..., N\}$. 
Then we have
\begin{equation}
\hat{y}_i = softmax(W_i^{q}*r_i+b_i^{q})
\end{equation}
Where $W_i^{q} \in \mathbb{R}^{C*d}$ and $b_i^{q} \in \mathbb{R}^{C}$ are trainable parameters of the softmax layer. 
$C$ is the number of labels (i.e, $3$ in our task).
Hence, the ACSA loss for a given review $R$ is defined as follows,
\begin{equation}
    loss_{ACSA} = \frac{1}{K}\sum_{i=1}^N p_i \sum_{C}y_i*\log{\hat{y}_i}
\end{equation}
If the aspect category $a_i$ is not mentioned in $S$, $y_i$ is set as a random value. 
The $p_i$ serves as a gate function, which filters out the random $y_i$ and ensures only the mentioned aspect categories can participate in the calculation of the loss function.

\noindent\textbf{Rating Prediction}
Since the objective of RP is to predict the review rating based on the review content, we adopt the [CLS] embedding $h_{[cls]} \in \mathbb{R}^d$ BERT produces  as the representation of the input review, where $d$ is the size of hidden layers in the BERT encoder. 
\begin{equation}
    \hat{g} =\beta^T* \tanh(W^r*h_{[cls]} + b^r)
\end{equation}
Hence the RP loss for a given review $R$ is defined as follows,
\begin{equation}
    loss_{RP} = |g - \hat{g}|
\end{equation}
Where $W^{r} \in \mathbb{R}^{d*d}$,$b^{r} \in \mathbb{R}^{d}$, $\beta \in \mathbb{R}^d$ are trainable parameters.

The final loss of our joint model becomes as follows.
\begin{equation}
    loss = loss_{ACSA} + loss_{RP}
\end{equation}

%% file: 5-Experiments.tex
\section{Experiments}
\label{sec:experiment}
We perform an extensive set of experiments to evaluate the performance of our joint model on \ABC{} and \textsc{Restaurant}~\citep{SemEval2014}.
Ablation studies are also conducted to probe the interactive influence between ACSA and RP.
\subsection{ACSA}
\begin{table*}[htp]
	\small
	\centering
	\caption{The experimental results of ACSA models on \textsc{ASAP} and \textsc{Restaurant}. Best scores are boldfaced. }\label{tb:exp-result}
	\begin{tabular}{llcccc}
	\hline
		 \multirow{2}{*}{Category}&  \multirow{2}{*}{Model}  & \multicolumn{2}{c}{\textsc{ASAP}}  & \multicolumn{2}{c}{\textsc{Restaurant}} \\ \cline{3-6}
		 &  & Macro-F1 & Acc. & Macro-F1 & Acc. \\ \hline
		\multirow{4}{*}{Non-BERT-based models} & TextCNN~\citep{kim2014convolutional} &  $60.41\%$ & $71.10\%$ & $70.56\%$ & $82.29\%$ \\
		& BiLSTM+Attn~\citep{zhou2016attention} & $70.53\%$ & $77.78\%$ & $70.85\%$ & $81.97\%$ \\
		& ATAE\_LSTM~\citep{ATAE} & $76.60\%$ & $81.94\%$ & $70.15\%$ & $82.12\%$ \\
		& CapsNet~\citep{capsnet} & $75.54\%$ & $81.66\%$ & $71.84\%$ & $82.63\%$ \\ \hline
		\multirow{3}{*}{BERT-based models} & Vanilla-BERT~\citep{Devlin2018BERTPO} & $79.18\%$ & $84.09\%$ & $79.22\%$ & $87.63\%$  \\
		& QA-BERT~\citep{sun2019utilizing} & $79.44\%$ & $83.92\%$ & $80.89\%$ & $88.89\%$ \\
		& CapsNet-BERT~\citep{mams} & $78.92\%$ & $83.74\%$ & $80.94\%$ & $89.00\%$ \\
		& Joint Model (w/o RP) & $80.75\%$ & $85.15\%$ & $\mathbf{82.01\%}$ & $\mathbf{89.62\%}$ \\
		& Joint Model & $\mathbf{80.78\%}$ & $\mathbf{85.19\%}$ & - & - \\ \hline
	\end{tabular}
\end{table*}
\noindent \textbf{Baseline Models}
We implement several ACSA baselines for comparison. 
According to the different structures of their encoders, these models are classified into Non-BERT based models or BERT-based models.
Non-BERT based models include TextCNN~\citep{kim2014convolutional}, BiLSTM+Attn~\citep{zhou2016attention}, ATAE-LSTM~\citep{ATAE} and CapsNet~\citep{capsnet}.
BERT-based models include vanilla BERT~\citep{Devlin2018BERTPO}, QA-BERT~\citep{sun2019utilizing} and CapsNet-BERT~\citep{mams}.

\noindent \textbf{Implementation Details of Experimental Models }
In terms of non-BERT-based models, we initialize their inputs with pre-trained embeddings.
For Chinese \textsc{ASAP}, we utilize Jieba\footnote{\url{https://github.com/fxsjy/jieba}} to segment Chinese texts and adopt Tencent Chinese word embeddings~\cite{song2018directional} composed of $8,000,000$ words.
For English \textsc{Restaurant}, we adopt $300$-dimensional word embeddings pre-trained by Glove~\cite{pennington2014glove}.

In terms of BERT-based models, we adopt the $12$-layer Google BERT Base\footnote{\url{https://github.com/google-research/bert}} to encode the inputs.

The batch sizes are set as $32$ and $16$ for non-BERT-based models and BERT-based models respectively. 
Adam optimizer~\cite{kingma2014adam} is employed with $\beta_1 = 0.9$ and $\beta_2 = 0.999$. 
The maximum sequence length is set as $512$.
The number of epochs is set as $3$.
The learning rates are set as $0.001$ and $0.00005$ for non-BERT-based models and BERT-based models respectively. 
All the models are trained on a single NVIDIA Tesla $32$G V$100$ Volta GPU.

\noindent \textbf{Evaluation Metrics}
Following the settings of \textsc{Restaurant}, we adopt Macro-F1 and Accuracy (Acc) as evaluation metrics.

\noindent \textbf{Experimental Results \& Analysis}
We report the performance of aforementioned models on \textsc{ASAP} and \textsc{Restaurant} in \autoref{tb:exp-result}.
Generally, BERT-based models outperform Non-BERT based models on both datasets.
The two variants of our joint model perform better than vanilla-BERT, QA-BERT, and CapsNet-BERT, which proves the advantages of our joint learning model.
Given a user review, vanilla-BERT, QA-BERT, and CapsNet-BERT treat the pre-defined aspect categories independently, while our joint model combines them together with a multi-task learning framework.
On one hand, the encoder-sharing setting enables knowledge transferring among different aspect categories.
On the other hand, our joint model is more efficient than other competitors, especially when the number of aspect categories is large.
The ablation of RP (i.e., joint model(w/o RP)) still outperforms all other baselines.
The introduction of RP to ACSA brings marginal improvement. 
This is reasonable considering that the essential objective of RP is to estimate the overall sentiment polarity instead of fine-grained sentiment polarities.

\begin{figure}[htb]
\small
	\centering
	\includegraphics[height=0.4\textwidth]{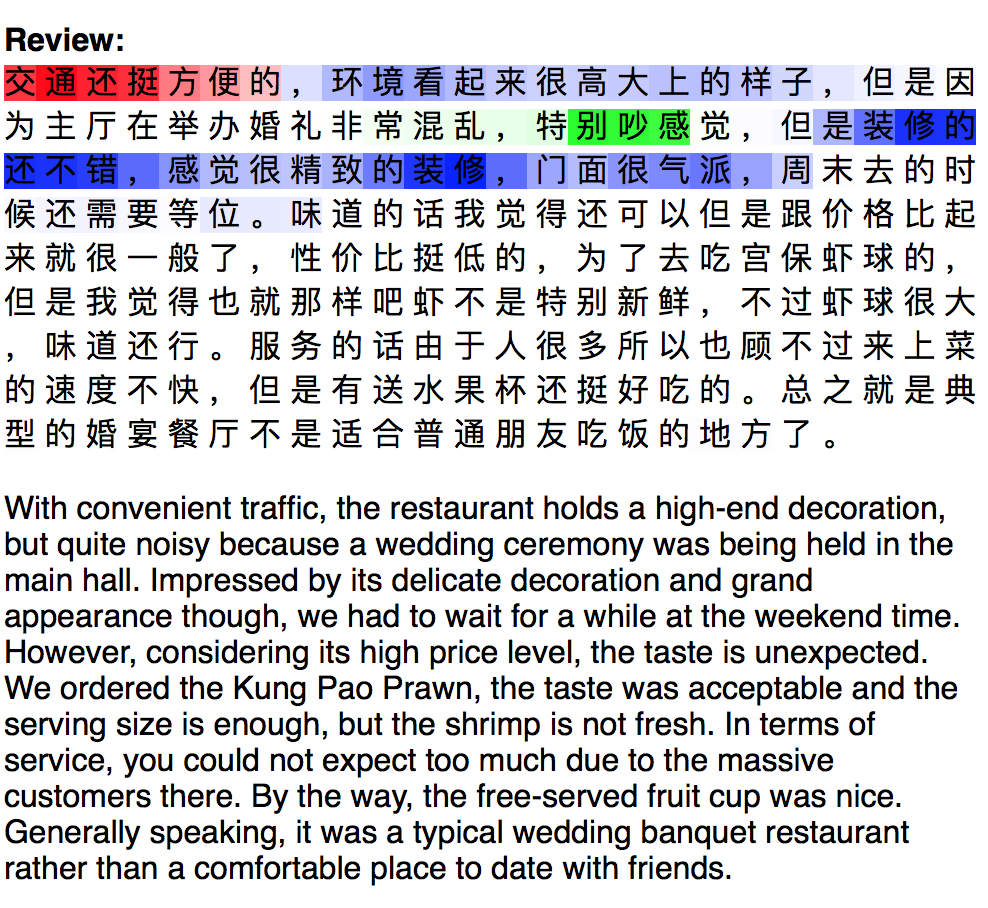}
	\caption{Attention visualization example. We only show attention weights of $3$ aspect categories for beauty. The red text span "..With convenient traffic..(\begin{CJK*}{UTF8}{gkai}..交通还挺方便的..\end{CJK*})" is related to \textit{Location\#Transportation}. The blue text span "..the restaurant holds a high-end decoration..Impressed by its delicate decoration and grand appearance though..(\begin{CJK*}{UTF8}{gkai}..环境看起来高大上的样子..装修还不错，很精致的装修..\end{CJK*})" is related to \textit{Ambience\#Decoration}. The green text span "..but quite noisy..(\begin{CJK*}{UTF8}{gkai}..特别吵感觉..\end{CJK*})" is related to \textit{Ambience\#Noise}. The intensity of the color represents the magnitude of attention weight.} 
	\label{fig:attenion_score}
\end{figure}

We visualize the attention weights produced by our joint model on the example of \autoref{tb:dataset} in \autoref{fig:attenion_score}.
Since different aspect category information is dispersed across the review of $R$, we add an attention-pooling layer~\cite{ATAE} to aggregate the related token embeddings dynamically for every aspect category.
The attention-pooling layer helps the model focus on the tokens most related to the target aspect categories.
\autoref{fig:attenion_score} visualizes attention weights of $3$ given aspect categories.
The intensity of the color represents the magnitude of attention weight, which means the relatedness of tokens to the given aspect category. 
It's obvious that our joint model focus on the tokens most related to the aspect categories across the review of $R$.

\subsection{Rating Prediction}
We compare several RP models on \textsc{ASAP}, including TextCNN~\citep{kim2014convolutional}, BiLSTM+Attn~\citep{zhou2016attention} and ARP~\citep{arp}. 
The data pre-processing and implementation details are identical with ACSA experiments.

\noindent \textbf{Evaluation Metrics.} We adopt Mean Absolute Error (MAE) and Accuracy (by mapping the predicted rating score to the nearest category) as evaluation metrics.

\noindent \textbf{Experimental Results \& Analysis}
The experimental results of comparative RP models are illustrated in \autoref{tb:rating-result}.
\begin{table}[H]
\small
	\centering
	\caption{Experimental results of RP models on \ABC{}. Best scores are boldfaced.}\label{tb:rating-result}
	\begin{tabular}{p{0.27\textwidth}cc}
	\hline
		Model & MAE & Acc. \\ \hline
		 TextCNN~\citep{kim2014convolutional} &  $.5814$ & $52.99\%$ \\
		 BiLSTM+Attn~\citep{zhou2016attention} & $.5737$ & $54.38\%$  \\
		 ARP~\cite{arp} & $.5620$ & $54.76\%$ \\ 
		 Joint Model (w/o ACSA) & $.4421$ & $60.08\%$  \\
		 Joint Model & $\mathbf{.4266}$ & $\mathbf{61.26\%}$ \\ \hline
	\end{tabular}
\end{table}

Our joint model which combines ACSA and RP outperforms other models considerably. 
On one hand, the performance improvement is expected since our joint model is built upon BERT.
On the other hand, the ablation of ACSA (i.e., joint model(w/o ACSA)) brings performance degradation of RP on both metrics.
We can conclude that the fine-grained aspect category sentiment prediction of the review indeed helps the model predict its overall rating more accurately.

This section conducts preliminary experiments to evaluate classical ACSA and RP models on our proposed \textsc{ASAP} dataset.
We believe there still exists much room for improvements to both tasks, and we will leave them for future work.